\documentclass[lettersize,journal]{IEEEtran}
\usepackage{amsmath,amsfonts}
\usepackage{algorithmic}
\usepackage{algorithm}
\usepackage{array}
\usepackage[caption=false,font=normalsize,labelfont=sf,textfont=sf]{subfig}
\usepackage{textcomp}
\usepackage{stfloats}
\usepackage{url}
\usepackage{verbatim}
\usepackage{graphicx}
\usepackage{cite}
\begin{document}

\title{ Toward Real Flare Removal: A Comprehensive Pipeline and A New Benchmark}

\author{Zheyan Jin, Shiqi Chen, Huajun Feng, Zhihai Xu, Yueting Chen
\thanks{This paper was produced by the IEEE Publication Technology Group. They are in Piscataway, NJ.}
\thanks{Manuscript received April 19, 2021; revised August 16, 2021.}}

\markboth{Journal of \LaTeX\ Class Files,~Vol.~14, No.~8, August~2021}%
{Shell \MakeLowercase{\textit{et al.}}: A Sample Article Using IEEEtran.cls for IEEE Journals}

\IEEEpubid{ }

\maketitle

\begin{abstract}
Photographing in the under-illuminated scenes, the presence of complex light sources often leave strong flare artifacts in images, where the intensity, the spectrum, the reflection, and the aberration altogether contribute the deterioration. Besides the image quality, it also influence the performance of down-stream visual applications. Thus, removing the lens flare and ghosts is a challenge issue especially in low-light environment.
However, existing methods for flare removal mainly restricted to the problems of inadequate simulation and real-world capture, where the categories of scattered flares are singular and the reflected ghosts are unavailable. Therefore, a comprehensive deterioration procedure is crucial for constructing the dataset of flare removal.
Based on the theoretical analysis and real-world evaluation, we propose a well-developed methodology for generating the data-pairs with flare deterioration. The procedure is comprehensive, where the similarity of scattered flares and the symmetric effect of reflected ghosts are realized. 
Moreover, we also construct a real-shot pipeline that respectively processes the effects of scattering and reflective flares, aiming to directly generate the data for end-to-end methods.
Experimental results show that the proposed methodology add diversity to the existing flare datasets and construct a comprehensive mapping procedure for flare data pairs. And our method facilities the data-driven model to realize better restoration in flare images and proposes a better evaluation system based on real shots, resulting promote progress in the area of real flare removal.

\end{abstract}


\begin{figure}
  \centerline{\includegraphics[width=0.5\textwidth]{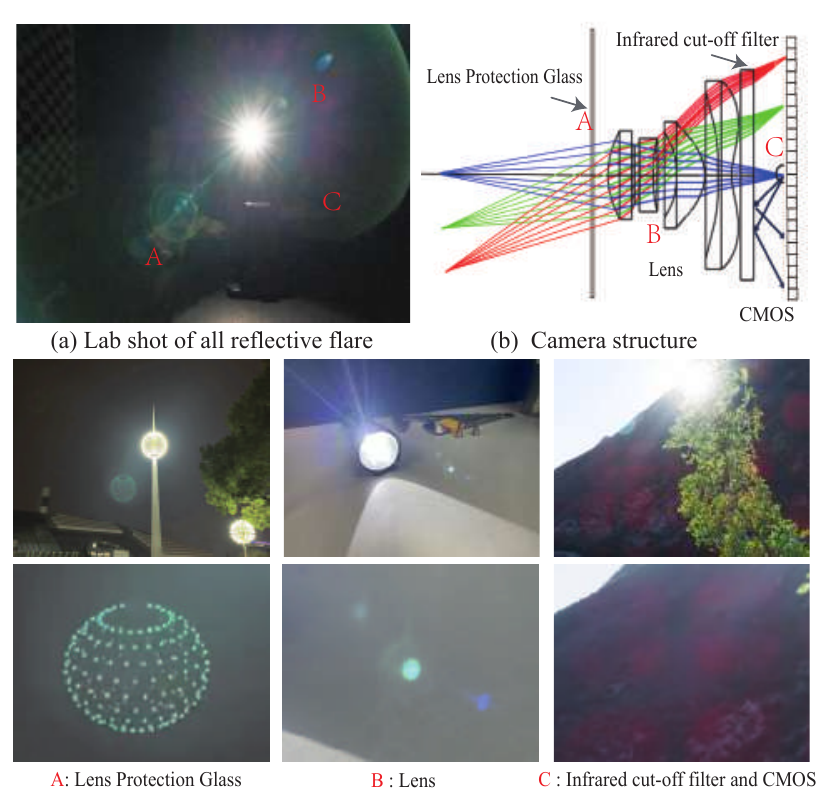}}

  \caption{\textbf{Various types of flare and where it occurs.}  (a) An image with mostly flare types is taken under special lab conditions.  (b) Schematic diagram of the location of glare in the lens structure. (c) Different types of reflective flare illustrated. The ABC corresponds to the previous images.}
  \label{fig1}
\end{figure}
\section{Introduction}

\IEEEPARstart{P}{hotographs} of scenes with a strong light source often exhibit lens flare, which is a salient visual artifact caused by unintended reflections and scattering within the camera. Flare artifacts can be distracting, reduce detail, and occlude image content. Although great efforts have been made in optical design to minimize lens glare, both consumer cameras and expensive professional film lenses can produce a large amount of flare when exposed to unsuitable light sources.

The lens flare pattern depends on the optical characteristics of the lens, lens protection glass, infrared cut-off glass, position and intensity of the light source, manufacturing defects, scratches and dust accumulated through daily use. The diversity of causes of lens flare leads to a variety of manifestations. Typical artifacts include halos, streaks, bright lines, saturated spots, reflections, haze and so on. This diversity makes removing glare a challenging problem.

Most existing methods for generating lens flares do not take into account the physical factors that cause flares, but simply rely on flare templates and intensity threshold to synthesize data. Flare templates are often generated from laboratory shots, so they can only detect and potentially remove a limited number of flare types and do not work well in more complex real-world scenarios.

Despite the emergence of various data generation methods in the field of glare removal, the main challenge remains a lack of training data. Collecting data requires outdoor nighttime static scenes, imaging positions that are completely unchanged, and constant switching of imaging devices. Existing methods achieve no-glare and glare images by switching lenses \cite{flare7K}. This method requires post-registration and is difficult to maintain scene illumination. Other researchers have manually placed an occluder between the light source and the camera \cite{2020How}, but this method is too labor-intensive and difficult to compensate for the central light source. Due to the complex causes of lens flare, even changing lenses or blocking light sources cannot guarantee that new images obtained are completely free of glare.

To overcome this challenge, we propose to generate real-world datasets using a purely optical approach. We generate simulated flare using a purely optical approach and create simulated datasets based on prior knowledge of optical physics.
 
For scattering situations (e.g. scratches, dust, oil and other defects), we propose the most reliable real-world data collection method that does not require registration, is fast to collect, preserves central light sources, and is not limited by camera lens parameters. Secondly, we also construct an optical model that demonstrates the existence of a synergistic effect between scattered lens flares through theoretical derivation and device experiments. Finally, we also believe that the number and displacement of light sources in each image in the real world may vary, with multiple light sources being possible. We perform multi-light source data augmentation on simulated data for these improvements. We have demonstrated these improvements through datasets and network experiments.

For reflection cases (e.g., lens reflections, protective glass reflections), we propose a method to remove the reflection flare without changing the lens. Based on real-world experiments and optical lens model analysis, we find that reflection flares are often symmetrical to the center of the main light source. Furthermore, when the camera shifts, the position of the reflective flare often differs from that of the normal subject. Based on 3D reconstruction and super-resolution methods, we can directly obtain a reconstructed image without reflective flare by inputting an image with reflective flare. We also apply this center symmetry effect of reflective flare to simulated data construction. We have discussed the existence of a center symmetry effect of glare through theoretical derivation and image analysis. We have also demonstrated the effectiveness of this improvement through datasets and network experiments.

To demonstrate the effectiveness of our new real-world dataset, data augmentation, synergistic effect, and center symmetry effect, we trained two different neural networks \cite{uformer,restormer} that were originally designed for other tasks. Both subjective and objective effects proved the effectiveness of our method. By using our real-world dataset to evaluate network performance, our data had a significant effect on both subjective and objective evaluations of network ability.

\begin{figure*}
	\centering
    \centerline{\includegraphics[width=1\textwidth]{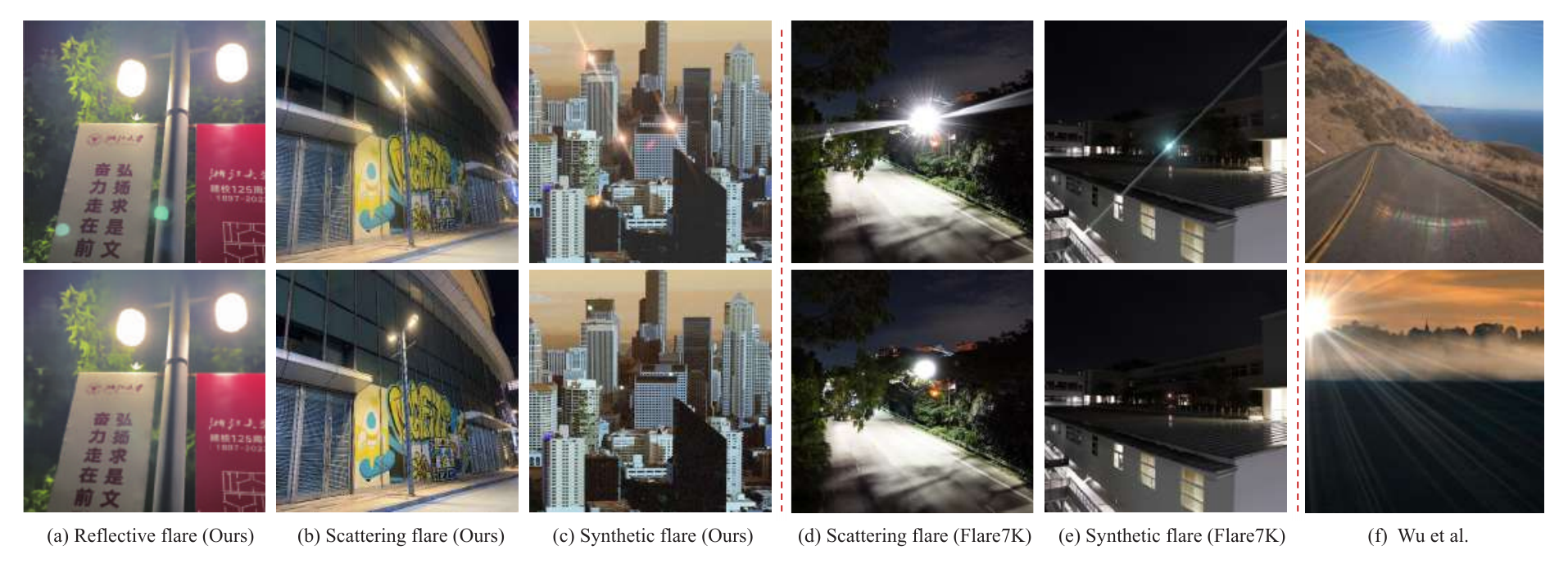}}
    \caption{\textbf{Comparison of different deflare datasets.} The existing real-shot dataset (f) does not have a suitable ground truth. Neither the real-shot dataset (d) nor the simulation dataset (e) could well restore the color and shape of the light source, and the flare does not conform to the real world. The simulation dataset (c) we proposed considers the multi-light source situation well and also takes into account the similar scattering flare caused by different position light sources for the same lens, and adds centrally symmetric reflected glare. Our real-shot dataset can capture complex multi-light source scenes well and can accurately make data on reflection-type flare. }
    \label{fig2}
    \end{figure*} 

\section{Background knowledge and previous art}
\subsection{Single Image Flare Removal}

Hardware solutions for glare usually involve complex optical designs and materials to reduce flares. Anti-reflective coatings are applied to lens components to reduce internal reflections. This coating can only be optimized for specific bands. Since the working length and incident angle of the lens are constantly changing, coating cannot be perfect. Adding a coating to an optical surface is expensive, so many post-processing techniques have been proposed to remove flares.
 
Many digital processing methods are used for flares in HDR photography \cite{H19}. These methods depend largely on the assumption of a point spread function (PSF). However, the PSF varies spatially, and the assumptions of the article are not true. Other solutions \cite{H1}\cite{H24} use a two-stage process: detecting flares based on the unique shape, position, and image degradation caused by lens flares. These solutions only apply to limited types of flares and are easily misclassified as all bright areas being flares. Existing methods often remove all glare areas, but images contaminated by flares are more like a local haze, and the perception and recovery of data information under the haze are the key points.

Some learning-based methods for flare removal \cite{2020How}, \cite{F16}, \cite{F22}, \cite{F5} have been proposed. These methods can be used for removing flares during the day or removing flares with specific pattern types.

Recent works have applied methods from other tasks to single-frame flare removal, such as dehazing \cite{dehazeformer}, removing reflections \cite{H15} and denoising \cite{H1}. These methods attempt to decompose an image into normal (ground truth) and glare parts, essentially splitting the image into two layers. The performance of this type of network largely depends on a high-quality training dataset. There have been many works focusing on generating flare datasets, but there are still many areas that need improvement.

\subsection{Flare Removal Datasets}

In the method proposed by Wu et al \cite{2020How}, physical solar flare images are captured in a darkroom and overlaid on the clear images to create paired data. Sun et al. \cite{F22} assumed that all flares are produced with the same point spread function (PSF), resulting in relatively uniform generated flares. With the help of Optical Flares (a plugin used for rendering lens flares in Adobe) in Adobe After Effects, Flare7K \cite{flare7K} can render lens flares. They developed a large-scale hierarchical flare dataset, which provides 25 types of scattering flares and 10 types of reflective flares that can be used to generate data. However, these datasets have many limitations and shortcomings:

1  
Lack of reliable real-world datasets: The few available datasets for scattering flare testing are often obtained by switching between different lenses, requiring post-registration of these images. The poor reliability of this approach, coupled with its high cost, has resulted in a limited number of real-world scattering flare removal datasets. Additionally, there is also a shortage of real-world reflective flare datasets.

2
Limited range of lens types: Many of these studies acknowledge that glare data is highly dependent on the lens, but they are only able to collect data from a single type of lens. This limits their ability to quickly gather glare data from various types of lenses.

3
There is a lack of in-depth understanding on the causes of glare. 
Specifically, there is difficulty in accurately differentiating between scattered glare and reflection glare. In reality, scattered glare produces global scattering rather than just simple speckles. Reflection glare can occur not only inside the lens but also on the lens protection glass and CMOS protection glass. Past datasets have not considered these factors sufficiently.

Secondly, it is a mistake to assume that more professional lenses do not produce flare. In fact, fingerprints, dust, wear and tear can cause diffraction and reflection in all lenses. Although professional lenses have more advanced optical coatings, larger apertures, lenses, and sensors often produce different types of flare.

\subsection{NeRF-based View Rendering}
Neural Radiance Fields (NeRF) ~\cite{NeRF} is an implicit MLP-based model that maps 5D vectors—3D coordinates plus 2D viewing directions—to opacity and color values, computed by fitting the model to a set of training views.
NeRF++ ~\cite{2020NeRF}tries to solve the ambiguity problem of image reconstruction in NeRF, and present a novel spatial parameterization scheme. PixelNeRF ~\cite{pixelNeRF} can be achieved with fewer images. NGP ~\cite{NGP} use of hash coding and other acceleration methods greatly speeds up the computation of neural radiation fields. 

In addition to the reduction in the number of images required and calculation overhead, there are variations of NeRF type algorithms for each area. BlockNeRF~\cite{BlockNeRF} which focuses on street view generation, megaNeRF~\cite{MegaNeRF} which focuses on large scale images, wildNeRF~\cite{wildNeRF} which focuses on image fusion in different exposure ranges, and darkNeRF~\cite{darkNeRF} which focuses on low illumination at night.

\section{Physics of lens flare}
In the ideal lens design, a clump of rays should converge on the sensor with normal operating path. However, the presence of complex light source will enlarge the defects of manufactured cameras, where the scattered and reflected light propagates along unintended paths to the image plane. These problems are common in the real-world scenes with high dynamic range, especially in nighttime or under the sun.
In this section, we first explain the physical mechanism of different flares, where they are mainly divided into two categories: scattering and reflecting flare. Then the challenges of existing single-image glare removal are introduced.

\begin{figure*}
	\centering
    \centerline{\includegraphics[width=1\textwidth]{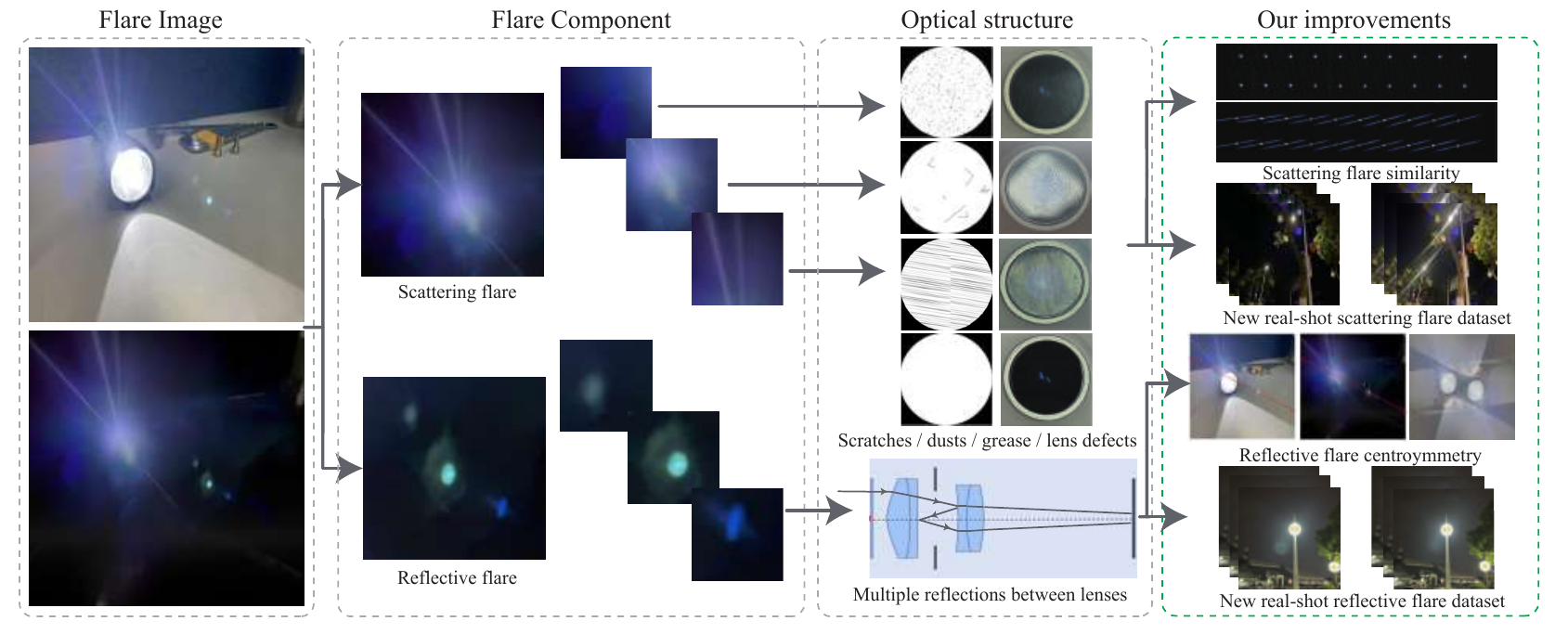}}
    \caption{\textbf{Overview of flare causes and our improvements.} Flare can be divided into scattering flare and reflective flare. Each type flare can be further subdivided. Each type of flare has its corresponding optical degradation reasons. We have proposed simulation and real shot dataset improvements for scattering and reflective flare separately. }
    \label{fig3}
    \end{figure*}  

\subsection{Scattering flare}
The dust, scratches, and grease in manufactured lenses will introduce diffraction to the incident light, where rays are scattered or diffracted not along their intended path, causing degradation. Dust adds an iridescent effect, while scratches bring one or more streaks of rays from the light source as a starting point. Scattering may also reduce the contrast in the area around the light source.

\subsection{Reflective flare}

In a practical lens system, a small amount of reflection occurs at each air-glass interval. After an even number of reflections, the light propagation direction changes slightly and sometimes still reaches the image plane. Because the number of air-glass intervals is twice that of the lens, there are more opportunities to form reflection flares.

In the spatial distribution of reflective flare, the optical path is symmetric because of the circular aperture. In the image, these reflective flares are usually located on a straight line connecting the light source and the center point. They are sensitive to the angle of incidence of the light source. The distribution of the reflective flares is related to the field of view. The reflective flare is also limited by the mechanical structure of the lens such as the aperture. Part of the structure can block the light path of the reflected flare, leading to arcing artifacts.

In the spectrum distribution, reflection flare also behaves differently on different spectra. Due to the different lens transmittance of different wavelengths of light, the camera lens suffers from chromatic aberration. At the same time, in order to reduce the reflection of air glass spacing and improve the transmittance, the lens surface is commonly coated with optical coating. However, this coating responds differently to different spectral bands. So the color of the reflected flare varies a lot and is difficult to simulate.


Meanwhile, there are two commonly overlooked situations of reflective glare: protection glass reflection and CMOS reflection caused by infrared cut-off glass. Protection glass reflection often produces flare that is symmetrical to the light source and loses image information due to overexposure. Reflective flare often exhibits more complex textural changes in light sources, providing new ideas for recovering image information of light sources. The reflective flare caused by the infrared cut-off glass CMOS is created when light hits the CMOS and is diffracted and reflected back onto the CMOS by the periodic circuit structure on the CMOS, which then bounces off the infrared cut-off glass. Both types of flare are not only related to the lens but also to the entire imaging device, making data collection difficult and, currently, there are no viable solutions.
 
\subsection{Challenges in flare removal}
Flares vary based on the location, size, intensity and spectrum of the light source, as well as the design and imperfections of the lens. Therefore, multiple flares can also appear in the same image. In order to deal with possible complex situations, it is almost impossible to establish a complete algorithm to analyze, identify and eliminate each type of flare. Therefore, we focus on a data-driven approach to solving complex flare, where high-quality and accurate data is most important for the popular deep learning methods.

\section{Proposed Method}
\subsection{Scattering flare similarity}

\begin{figure*}
	\centering
    \centerline{\includegraphics[width=1\textwidth]{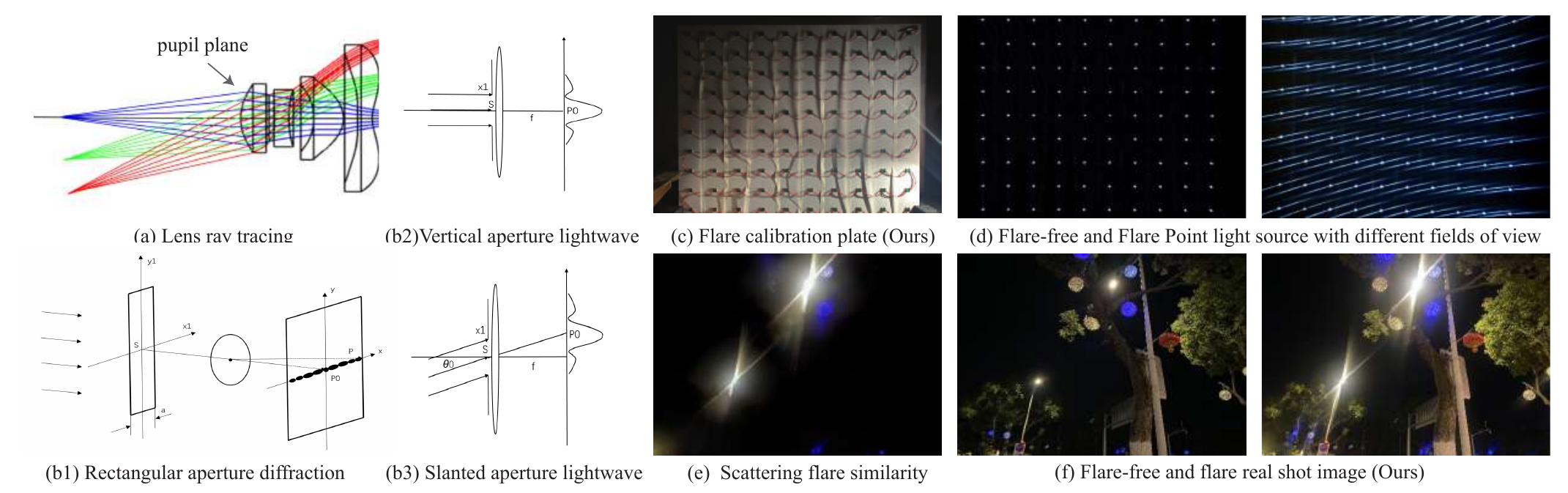}}
    \caption{\textbf{Similarity effects of scattering flare.} The aperture plane of a typical consumer camera is located at the first surface of the lens, as shown in (a). The light entering from different angles will be affected by the same contaminated incident aperture. (b1) represents rectangular optical diffraction, (b2) and (b3) represent diffraction results after oblique and vertical incidence. Different light sources in different fields of view usually produce similar flare results at the same aperture plane and slightly different angles, as shown in (d). (c) shows the experimental setup of (d). (e) shows the similarity of scattering flare from different light sources in real captured images. (f) shows the data pairs of the image in (e). }
    \label{fig4}
\end{figure*}

The pupil plane of a typical consumer-level camera generally locates on the first surface of lens, where the light entering from different angles will be equally affected by the polluted entrance pupil. Therefore, the light sources of different fields of views (FoVs) produce similar flare results, as shown in Fig. \ref{fig4}(d).
When light passes through an imperfect lens, the far-field Fraunhofer diffraction occurs, which can be formulated as follows, 

\begin{equation}
\begin{split}
E = & C \frac{A^{\prime}}{f} \exp (i k f) \exp \left[i k\left(\frac{x^2+y^2}{2 f}\right)\right]   \\
& \iint_{\Sigma} \tilde{E}\left(x_1, y_1\right) \exp \left[-\mathrm{i} k\left(l x_1+\omega y_1\right)\right] \mathrm{d} x_1 \mathrm{~d} y_1,
\end{split}
\label{flhf}
\end{equation}
here, the term $C \frac{A^{\prime}}{f} \exp (i k f)$ represents the phase delay from S to P, where S and P are indicated in Fig. \ref{fig4}(b1-b3). $\tilde{E}\left(x_1, y_1\right)$ represents the intensity distribution of the incident light on the diffraction plane, which is the same as the intensity distribution on the aperture of a camera. $\exp \left[-\mathrm{i} k\left(l x_1+\omega y_1\right)\right]$ represents the phase difference between each point relative to the center point of the entrance pupil. Based on this formula, the structure of the first surface performs as a physical Fourier transform operator to generate the scattering flare.

A damaged lens that causes scattering is equivalent to a complex diffraction element. In one scene, there are various light sources of different positions and intensities. The complex amplitude from different FoVs can be represented as,

\begin{equation}
\label{E0}
\small
E_\theta=E_0 \exp \left[i 2 \pi x_1 \frac{\sin \theta_0}{\lambda}\right],
\end{equation}
where $\theta_0$ represents the angle between the incident direction and the optical axis of camera. $\lambda$ represents the wavelength of light. This representation of complex amplitude forms a plane wave inclined to the diffraction aperture plane (spatial domain). The spatial inclination is also equivalent to a linear phase shift in Fourier domain, which can also be expressed as,

\begin{equation}
\label{E0P}
\small
E_\theta=E_0 \exp \left[i 2 \pi x_1  u_0 \right],
\end{equation}
where the translation amount $u_0$ is expressed as:

\begin{equation}
\label{E0P}
\small
 u_0=\frac{x_0}{\lambda f}=\frac{\sin \theta_0}{\lambda},
\end{equation}
 
Due to the Fraunhofer diffraction of the incident light source requires Fourier transformation, while illuminating the aperture with a tilted plane wave will cause the shift to the diffraction results. This conclusion is also known as the phase shift theorem of Fourier transform, which states that there is a spatial displacement in the time domain and a phase shift in the frequency domain. The specific formula is given as follows:

\begin{equation}
\label{FPM}
\small
\mathcal{F}\left[f\left(x_1\right) \exp \left(i 2 \pi u_0 x_1\right)\right]=F\left(u-u_0\right),
\end{equation}
$\mathcal{F}$ represents the Fourier transform. The form of the frequency domain $F(u-u_{0})$ remains unchanged, only a horizontal shift occurs. The change in the central position still conforms to geometrical optics.

Based on the above formula derivation, the scattering caused by the same point light source in different image fields is similar. Due to the different colors and intensities of the light source, the size, brightness, and color of the scattering flare are slightly different, but the shape of the scattering image is similar and the spectrum is similar. To prove the similarity of different scattering flare from actual shooting, we built an experimental device that can fill the camera's full FoV with point light sources. 
We use various types of degraded lenses to capture scattering flare. As shown in Fig. \ref{fig4}(d)-(f), the similarity of scattering flare was maintained in the whole FoV. In summary, through theoretical derivation and real capture, we proved the similar effect of scattering flare. And this characteristic is applied to the generation of datasets, which is illustrated in the following.

\subsection{Scattering flare datasets}
For ordinary consumer lenses, the pupil plane is often the first surface of the camera lens. As long as the pupil plane is continuously destroyed and repaired, the desired flaring data of scattering can be obtained.

\begin{figure*}
    \centering
    \centerline{\includegraphics[width=1\textwidth]{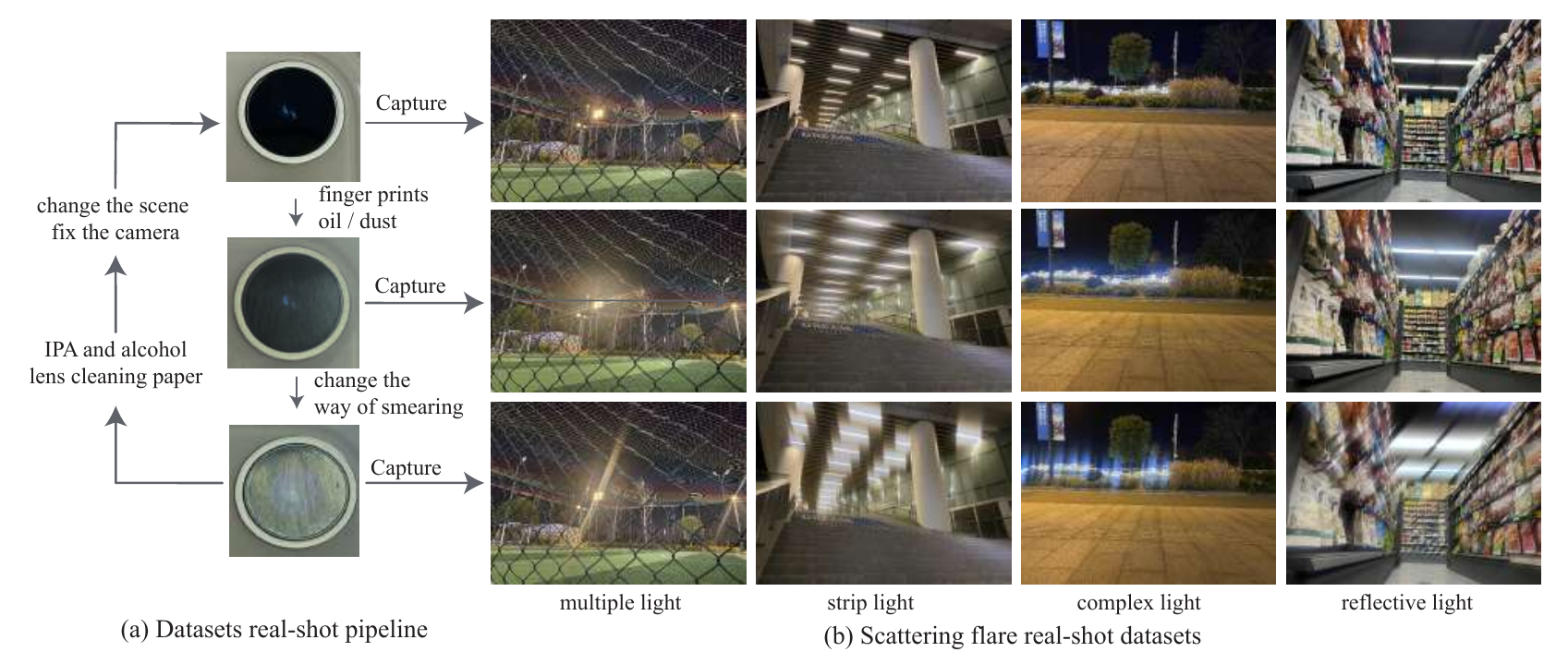}}
    \caption{\textbf{ Scattering flare real shot dataset. } (a) shows the actual shooting process and the change of the camera's protective glass (pupil surface). (b) shows various types of non-flare and flare images. Compared with existing real shooting datasets, our dataset pays more attention to the shape and quantity of light sources, as well as the reflection of objects. At the same time, in order to reduce the cost of shooting, we adopt a shooting process in which one non-flare image corresponds to multiple flare images.}
    \label{fig5}
\end{figure*}

Fig.\ref{fig5} shows the our process of capturing scattered flares. It mainly consists of the following steps:

\begin{itemize}
    \item [1)] Wipe the protective glass with Isopropylamine(IPA) and cleaning cloth.

    \item [2)] Find a suitable shooting position, where a light source in the scene will cause little flare, and hold on.

    \item [3)] Take the pictures of ground truth.

    \item [4)] Spread the oil and dust onto the protective glass.

    \item [5)] Take the pictures of flare.

    \item [6)] Repeat steps 4 and 5 continuously to obtain multiple groups of dazzling images and return to step 1.

\end{itemize}
Due to the influence of slight lens flare defects on the ground truth cannot completely avoid, the evaluation of paired real-world data can only serve as a reference. 
It cannot fully reflect the actual performance of the lens flare removal methods. However, compared to existing real shot datasets, our real scattered flare dataset has advantages such as a larger quantity of data, more diverse light source forms, more light sources, and the ground truth flare effect is smaller.

For the simulation dataset, we use the Optical Flares plugin in Adobe AE to generate data, just like Flare7K does. However, we improved the scattering synergistic effect and central symmetry effect in data generation. Compared to the flare of daylight, there are often multiple light sources in night scenes. We first generate multiple light sources of random size and position, and then generate corresponding flare of size and position based on the position and size of the light sources. The generated flare has a similar appearance, but the position and size are different, which conforms to the scattering flare synergistic effect in real world.

\subsection{Reflective flare centrosymmetry}
Reflective flare components are difficult to simulate by rendering techniques because they require an accurate description of the optics \cite{H10,H14}. However, the lens prescriptions are generally not available, which exacerbates the unpredictable of reflective flare. Fortunately, most existing optical lenses are mostly centrally symmetric, so no matter where the light source is in the camera, the flare data is distributed on the axis of symmetry of the light source. Moreover, because the lenses of similar design have proximate internal reflection. The data collected from one camera example can be used well for the lenses within the same application. 

Reflective flare has a lot to do with the structure of the light source, and the natural scenes are rich in lighting scenes. While in the laboratory, only the reflective flares from a single light source can be captured. Moreover, the structure of reflective flares varies at different FoVs. Thus, it is necessary to move the camera at various angles to capture the reflective flare components at different angles to ensure comprehensive data.


\subsection{Reflective flare datasets}

Factors such as imaging module, brightness and shape of light source, image background, and camera angle determine the reflection-type flare. Therefore, accurately simulating the reflection flare has a long way to go. However, the removal of reflection glare requires a large amount of reliable and diverse paired data.


It is difficult to simulate the components of reflective flares using rendering technology, because it requires an accurate description of optical elements, in which the precise prescription and the manufacturing deviations are hard to obtain \cite{H10,H14}. However, similar lenses have similar internal reflection paths, so data collected from one camera instance can usually be well generalized to other instances that use the same lens. For example, data generated using one consumer-grade camera can be applied to all the devices within the same scope.

Flare7k can simulate a dynamic reflective flare effect. The flare component's opacity is set to be proportional to the distance between the aperture and the light source. In order to demonstrate the clipping effect of the aperture on reflective flare, when the distance between the reflective glare component and the clipping threshold is greater than the threshold, the algorithm will erase part of the flare. The advantage of simulated data is its low cost and large quantity.

Based on the central symmetry effect, we have also improved the existing simulation method for reflective flares. The existing reflective glare templates are generally based on laboratory experiments and have consistent directions, such as the 45-degree oblique direction. We force the position of the diffuse flare and the light source to be placed on the central symmetry line of the reflective glare. Then we randomly rotate the entire glare template and fit it to the clean image.

Therefore, we propose a reflective flare removal method based on 3D reconstruction. When the camera's position changes, the light source and reflective flares in the image also change. Reflective flares are symmetrical with respect to the light source center. Different flares are generally distributed on the central axis of symmetry, and the displacement of reflective flares is also symmetrical with respect to the light source. In this time, the displacement changes of reflective flare do not conform to the laws of 3D reconstruction, so the reconstructed image will not have reflective flare.

The formula for 3D reconstruction is as follows:
\begin{equation}
 \boldsymbol{r}(t)=\boldsymbol{o}+t \boldsymbol{d},
\label{eqot}
\end{equation}
here $\boldsymbol{r}(t)$ is the sampled ray, $\boldsymbol{o}$ is the sampled ray emission point, $t$ is the displacement along the ray direction $\boldsymbol{d}$. Let $\sigma(\boldsymbol{r}(t))$ be the differential probability of the infinitesimal particle where the light ray terminates at position $t$. The color of this sampled ray is the integration from the near boundary $t_n$ to the far boundary $t_f$ is:

\begin{equation}
\boldsymbol{C}=(r, g, b)=\int_{t_{n}}^{t_{f}} T(t) \sigma(\boldsymbol{r}(t)) \boldsymbol{c}(\boldsymbol{r}(t), \boldsymbol{d}) d t,
\label{eqC}
\end{equation}

Where $\boldsymbol{c}$ represents the pixel value of the voxel at the corresponding position of $t$. $T(t)$ represents the cumulative transmission rate along the ray from $t_n$ to $t$, which is the probability that the ray propagates from $t_n$ to $t$ without being blocked:
\begin{equation}
T(t)=\exp \left(-\int_{t_{n}}^{t} \sigma(\boldsymbol{r}(s))ds\right),
\label{eqtx}
\end{equation}

\begin{figure*}
	\centering
    \centerline{\includegraphics[width=1\textwidth]{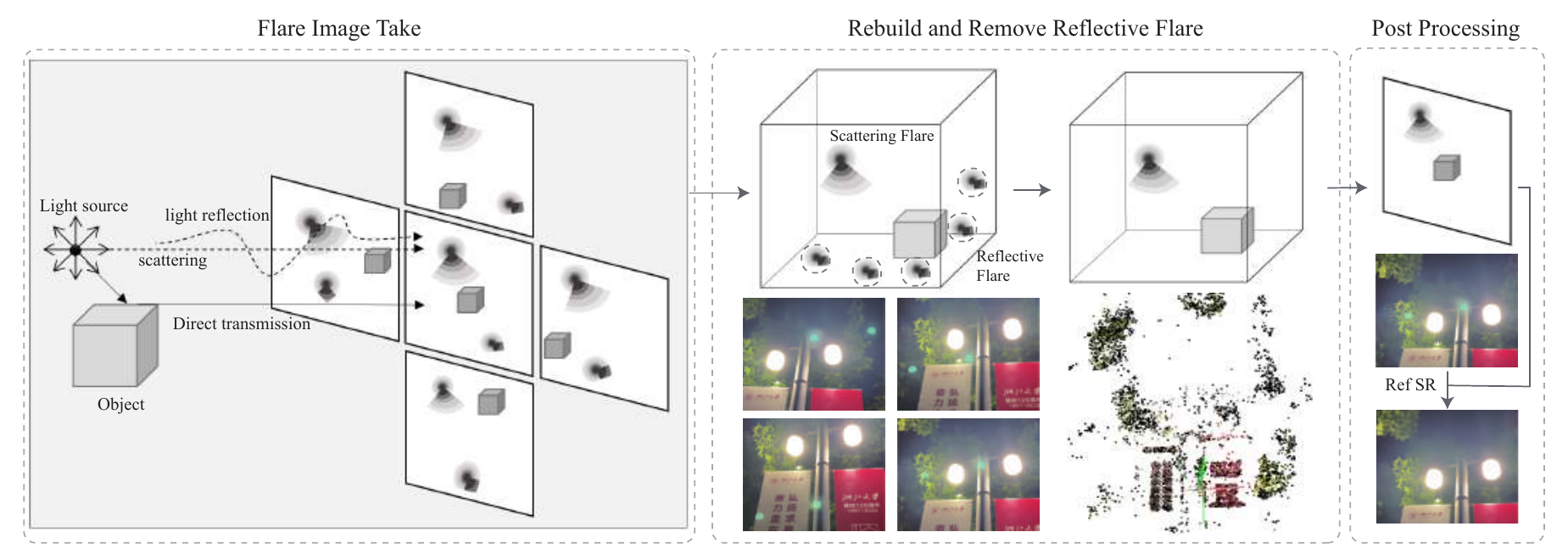}}
    \caption{\textbf{ Reflective flare real shot data set.}  (a) Shows images with reflective flare from different angles in the dataset. (b) Shows the optical flow information from different angles. Due to the central symmetry effect of reflective flare, the optical flow of reflective flare is significantly different from normal background. We can obtain high-quality data without any reflective flare through 3D reconstruction and the reference super-resolution method, such as (c) and (d).}
    \label{fig6}
    \end{figure*}

For a three-dimensional scene, objects and light sources are always stationary. Due to the center symmetry effect of reflective flare, when the camera's pose changes, the reflective flare varies a lot in spatial, which does not conform to the camera's pose change. As shown in Figure \ref{fig6} middle part, when all the information is input into the three-dimensional space, the reflective flare will appear at various positions in the space. The probability of occurrence at each position is the reciprocal of the entire dataset, and the sum of the probabilities of all positions is one. Since the flare occasionally blocks certain sampling rays, the color in this area must change. According to the derivation in the Nerf\cite{NeRF}, the integration can be discretized in the calculation process. We assume that the specular glare blocked at position $t_i$ to $t_{i-1}$, then the formula should be adjusted to:

\begin{equation}
\begin{aligned} 
\boldsymbol{C_i}=(r_i, g_i, b_i)=\int_{t_{n}}^{t_{i-1}} T(t) \sigma(\boldsymbol{r}(t)) \boldsymbol{c}(\boldsymbol{r}(t), \boldsymbol{d})dt \\
+\int_{t_{i-1}}^{t_{i}} T(t) \sigma(\boldsymbol{r}(t)) \boldsymbol{c}(\boldsymbol{r}(t), \boldsymbol{d})dt\\
+\int_{t_{i}}^{t_{f}} T(t) \sigma(\boldsymbol{r}(t)) \boldsymbol{c}(\boldsymbol{r}(t), \boldsymbol{d})dt,
\end{aligned} 
\label{eqCi}
\end{equation}

When reflective flare is added to the sampling ray from $t_i$ to $t_{i-1}$, it leads to changes in the transparency $\sigma(\boldsymbol{r}(t))$ and color $\boldsymbol{c}$ in the integral from $t_{i-1}$ to $t_i$. The color $\sigma(\boldsymbol{r}(t))$, transparency $\boldsymbol{c}$, and transmission rate $T(t)$ in the range from $t_n$ to $t_{i-1}$ have also changed.



The above formula is only the rendering process of the neural radiance field network. The actual network training only involves two types of parameters: $\sigma(\boldsymbol{r}(t))$ and $\boldsymbol{c}$. Since the flare $\boldsymbol{C_i}$ only appears once at the same position, the probability of $\boldsymbol{C_i}$ is much smaller than that of $\boldsymbol{C}$, when the network starts training and converges, the optimal solution will be close to a large number of flare-free $\sigma(\boldsymbol{r}(t))$ and $\boldsymbol{c}$. The erroneous parameters of the flare part will be automatically discarded to achieve the goal of removing the flare. It is worth noting that there are many improved versions of NeRF, which often use the image encoding feature to improve the performance of NeRF. However, these methods often do not perform well in fitting reflective glare, and the encoding feature will pay more attention to the interference of reflective flare.

Unlike capturing images of reflective flares in a laboratory, our data generation method can obtain datasets in any scene. This significantly improves the reliability and diversity of the data, reduces the difficulty of acquisition, and eliminates the need to consider complex issues related to the fusion of flare and images.

Compared to previous datasets, these designs better reflect real-world nighttime conditions.Three-dimensional reconstruction requires a large number of images from different angles, and the dataset for reflective flares also requires a large number of images from different angles. The requirements for both methods are perfectly matched.

In nighttime situations, various shapes and brightness-changing LED lights are common and may result in reflected light spots with various shapes and textures. Previously, there was no way to handle such situations, but our method will automatically handle these complex scenarios. Compared to previous datasets, these designs better reflect real-world nighttime conditions.


\subsection{Comparison with existing flare dataset}

\begin{table*}
  \centering
    \caption{ Comparison of the number and characteristics of different data sets in simulation and real shooting }
  
\begin{tabular}{lcccccccccc}
\hline { } & \multicolumn{5}{c}{ Simulation }  & \multicolumn{5}{c}{Real (Benchmark)}\\

Dataset & scattering & reflection  & type & similarity & centrosymmetry  & & scattering & reflection & light source & separation \\
\hline Wu et al.  & 2000 & 2000 & $2 + 1$   &  $\times$   & \checkmark & & 20 & 20  & $\times$ &   $\times$ \\
Flare7K & 7,000 & 7,000 & $25+10$ &   $\times$   &  $\times$  & & 100 & 0 &  \checkmark & $\times$ \\
Ours & 5000 * 5 & 5000 * 3 &  $25+10$ & \checkmark  &  \checkmark & & 500 & 2000 & $\checkmark$ & $\checkmark$ \\
  
   \hline
  \end{tabular}

  \label{tab:dataset}
\end{table*}

The benchmarks of flare dataset are proposed by Wu \cite{2020How} and flare7K \cite{flare7K}, which are designed for removing flare during daytime and nighttime. In the pipeline of \cite{2020How}, flare is mainly determined by the stains in front of the lens. Wu's method collects real flares by simulating all different pupil functions. However, the simulated pupil functions and real complex lens damage have a large discrepancy, leading to a diversity and accuracy gap between synthetic flares and real-world scenes.

The simulation part of the Flare7K dataset mainly uses AE simulation plug-ins, but does not take into account the physical causes of lens flare generation, such as flare similarity and centrosymmetry effects. Although the cost of simulation is low and the speed is fast, there are still noticeable differences between the simulation and the real flares.

At the same time, the Flare7K dataset does not have a real shooting dataset of reflective glare, which makes it impossible to measure the effect of simulated reflective glare.
 
Flare7K's scatter-type real-shot data is obtained using different cameras and lenses and then the images are registered, which results in higher costs and fewer quantity. In the contrast, our dataset generation has greater advantages of having both high-quality and large quantity of real-shot data sets at the same time. It can also obtain simulated data sets that are more in line with optical priors.

\section{Experimental Results}
\subsection{Simulation data test evaluation}


To demonstrate the effectiveness of our method, we trained two popular neural networks with different architectures that were previously used for other image restoration tasks. Using the proposed method, both networks produced satisfactory results. All network configurations were based on the most basic structure publicly available for each network.


\begin{figure*}
	\centering
    \centerline{\includegraphics[width=1\textwidth]{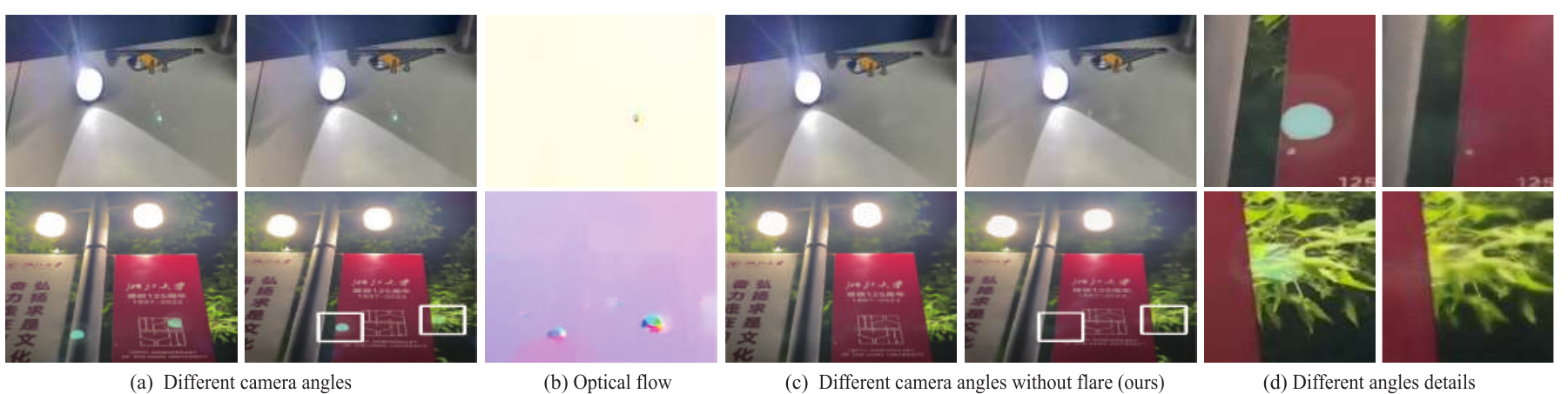}}
    \caption{\textbf{ Reflective flare real shot dataset.}  (a) Shows images with reflective flare from different angles in the dataset. (b) Shows the optical flow information from different angles. Due to the central symmetry effect of reflective flare, the optical flow of reflective flare is significantly different from normal background. We can obtain high-quality data without any reflective flare through 3D reconstruction and the reference super-resolution method, such as (c) and (d).}
    \label{fig7}
    \end{figure*} 

We trained Restormer \cite{restormer} and Uformer \cite{uformer} with our synthetic dataset. Clear images were obtained from the Flickr dataset \cite{H27}, while the scattered and reflected flare images were provided by Flare7K. The synthetic dataset included ten different variations:

\begin{itemize}
    \item base: complete same to Flare7K.

    \item R: with added reflective flare on base.

    \item RP: with added reflective flare based on centrosymmetric effect.

    \item MR. with added reflective flare and increased scattering flare.

    \item MRP. with added reflective flare based on centrosymmetric effect and increased scattering glare, and synergy effects.
    \item \textit{w/o} L. without light source. 
\end{itemize}

 
To ensure the fairness of evaluation, we used the same hardware and training time. The training dataset consisted of 5000 512*512 image pairs.
There are five major categories in the dataset, each of which is further divided into two categories based on whether there is a light source, amounting to a total of ten categories.


As shown in Table \ref{tab:train_ablation}, the performance of the two networks is similar on our dataset. Keeping the light source in GT preserves data that can train a better-performing network. Therefore, our later discussions are mainly based on the dataset that preserves the light source in GT. Simply added reflective flare may result in only a small improvement. However, increasing reflective flare based on the central symmetry effect can further enhance the network performance. Compared with dataset R, dataset MR increased scattering flare. Flare number augmentation can further improve the network's fitting ability. Compared with dataset RP, dataset MRP increased scattering flare based on the similarity effect, which has a much greater improvement than the data augmentation from dataset R to dataset MR. Compared with dataset MR, dataset MRP maintains the same amount of scattering flare, but introduces similarity effects, leading to improved training results. In summary, our experiments on the simulation dataset confirm that adding light sources in GT, increasing the number of scattering flare, adding reflective flare, based on similarity and central symmetry effects can all further improve network performance.

\begin{table*}
  \centering
  \caption{ Quantitative comparison of training with different synthetic types of data and testing on existing real flare datasets. }
  \begin{tabular}{lcccccccccc}
     \hline
 
Increase scattering flare  &  &    &   & \checkmark &  \checkmark    \\  
Reflective flare in input
  &           &  \checkmark &  \checkmark  & \checkmark     &  \checkmark   \\

Scattering flare similarity 
  &           &            &          &    & \checkmark  \\
Reflective flare centrosymmetry &       &      &  \checkmark     &  \checkmark  &     \checkmark     \\
 
\hline
Dataset no light source
  & base \textit{w/o}L  & R \textit{w/o}L & RP \textit{w/o}L & MR {w/o}L & MRP {w/o}L  \\
Restormer(PSNR)
  &24.04 (2.92\%) & 23.45 (10.2\%) & 24.15 (16.2\%) & \textbf{24.29} (0.00\%) & 23.91 (4.47\%) \\
Restormer(SSIM)
  &0.905 (4.40\%) & 0.898 (12.1\%) & \textbf{0.909} (0.00\%) & 0.907 (2.20\%) & 0.906 (3.30\%) \\
Uformer(PSNR)    
  & 24.39 (6.41\%) & 24.69  (2.80\%) & 23.13 (23.0\%)  & 24.47 (5.44\%) & \textbf{24.93} (0.00\%)  \\ 
Uformer(SSIM)    
   & 0.887 (24.2\%) & \textbf{0.909} (0.00\%) & 0.854 (60.4\%)   &   0.897 (13.2\%) &   0.902 (7.70\%)  \\ 
    \hline
Dataset Light source in GT & base & R & RP & MR & MRP \\

Restormer(PSNR) & 24.15 (12.7\%) & 24.38 (9.77\%) & 24.43 (9.14\%) & 24.45 (8.89\%) & \textbf{25.19} (0.00\%) \\

Restormer(SSIM) & 0.907 (5.68\%) & 0.902 (11.3\%) & \textbf{0.912} (0.00\%) & 0.909 (3.41\%) & \textbf{0.912} (0.00\%)\\

Uformer(PSNR) & 24.85 (2.21\%) & 24.88 (1.86\%) & 24.92 (1.39\%) & 24.99 (0.58\%) & \textbf{25.04} (0.00\%) \\

Uformer(SSIM)  & 0.910 (25.0\%) &   0.913 (20.8\%) & 0.918 (13.9\%) &  0.917 (15.3\%)  &  \textbf{0.928} (0.00\%) \\

   \hline
  \end{tabular}
  
  \label{tab:train_ablation}
\end{table*}

\subsection{Real shot data evaluation}

\begin{figure*}
	\centering
    \centerline{\includegraphics[width=1\textwidth]{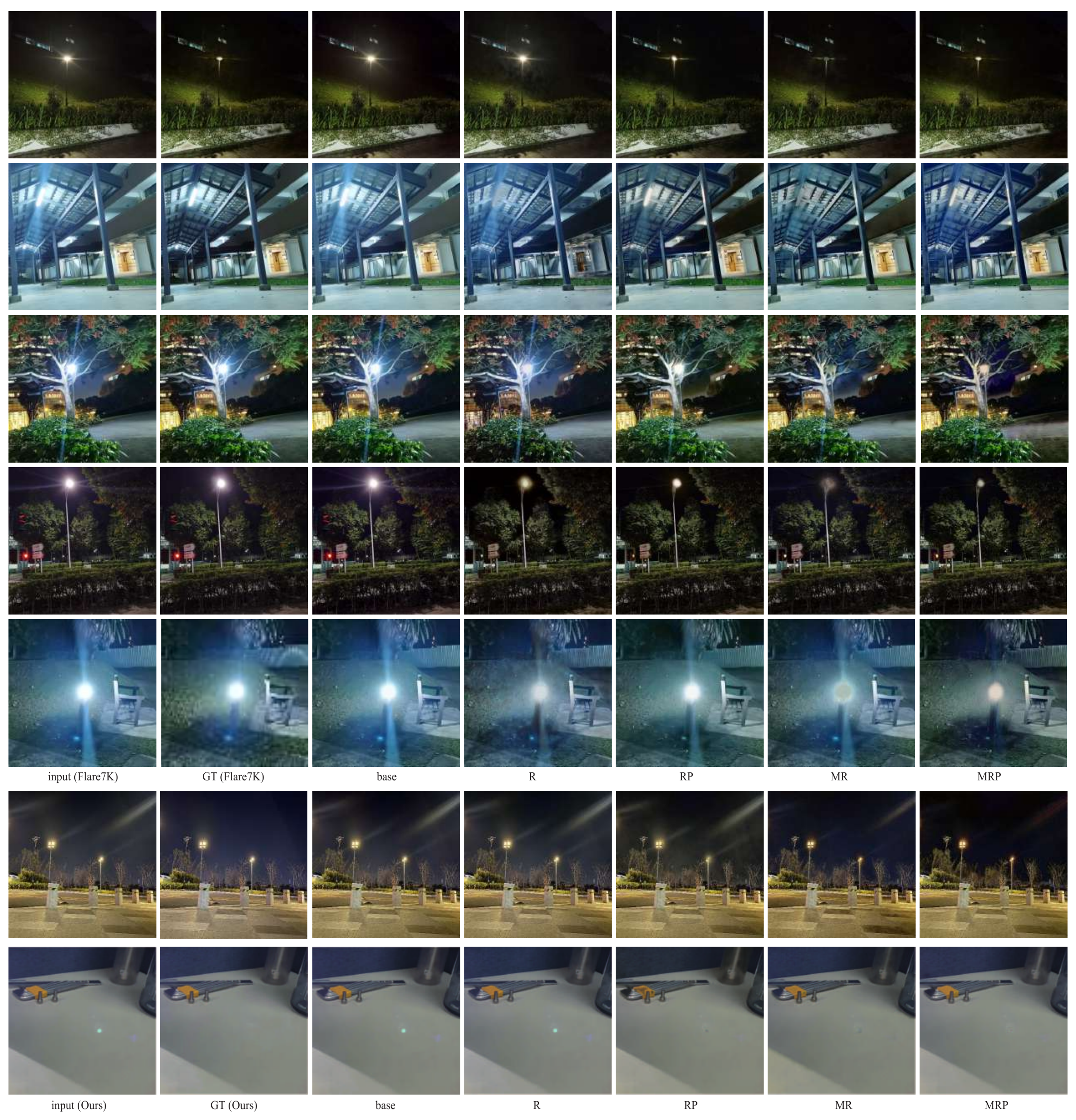}}
    \caption{\textbf{The test results of different datasets trained on the same network.} 
 The top half shows the results of flare7K's real captured data, while the bottom half shows our own real captured data. The impact of data improvement on the results can be observed from left to right. The advantages of our dataset compared to existing datasets are reflected from top to bottom. Our real shot image groundtruth scattering flares are smaller, light sources are more complex, and reflective flare is completely eliminated.}
    \label{fig8}
    \end{figure*} 

Due to flare area occupying an inconsistent part of the image area, many image indicators (PSNR, SSIM, LPIPS) can maintain high objective scores even without removing flare. Removing flare may cause slight degradation of non-flare areas and affect the overall image indicators. Therefore, it is essential to use various real shooting data for subjective testing comparison.

The upper half of Fig.\ref{fig8} shows the test results of flare7K real shooting data while the lower half shows the results of our own real shooting data of the same network training. From top to bottom, it demonstrates the advantages of our dataset over the existing flare7K dataset. Our real shooting data has smaller flares, more complex light sources, and can handle reflective flare. However, our real shooting dataset still has slight glare, which requires further efforts to address.


We have also discovered limitations of the flare7K dataset. Firstly, the flare in the provided ground truth is still obvious and not very convincing as a test benchmark. Moreover, as can be seen from the last line of flare7K part in Fig.\ref{fig8}, it not only has severe scattering flare but also fails to reflect the effects of reflective glare. The reflective flare in the ground truth is indistinguishable from that in the input image.

As shown in Fig.\ref{fig8}, from left to right, we can observe a significant improvement in subjective evaluation of network results due to changes in the training dataset. In the upper half of Fig.\ref{fig8}, we compare our results to the latest flare7K real shooting dataset. The experimental subjective results show complete consistency with our dataset improvement direction. Under the same training time, the existing methods like dataset base had relatively poor results and the flare removal effect was not significant. Dataset R added reflective flare, and the suppression ability was improved compared to dataset base, but the removal ability of scattering flare was not significantly enhanced. For example, in the bottom row of the flare7K part in the upper half of Fig.\ref{fig8} there is a blue reflective flare, and the removal of reflective flare in dataset base, dataset R, and dataset RP is getting better. Improving the number of flares had a poor effect on removing scattering flare. Dataset MR result may reduced the light source completely. Dataset MR also had little effect on reflective flare. However, dataset MRP, which combines all our data enhancement, can more effectively remove scattering flare and reflective flare. For example, as shown in the lower half of Fig.\ref{fig8}, the results of dataset MRP can remove most of the scattering flare while retaining the approximate shape of the light source. At the same time, the blue reflective flare in the last row can also be clearly suppressed.

The experimental results of dataset MRP show significant improvements in dealing with both scattering and reflective flare compared to the existing ground truth in real shooting datasets. This indicates that there is an urgent need to improve the testing real shooting datasets.

In the lower half of Fig.\ref{fig8}, we tested the network results using our own real shooting data. The training results of dataset base, dataset R, and dataset RP for removing scattering flare are average. Using dataset MR can significantly improve the flare removal result. After using the scattering flare similarity effect, dataset MRP has a very good result on removing scattering flare. Subjectively, the result of dataset MRP is even better than the ground truth image.
 
We generated our own real shooting reflective flare dataset, so we can subjectively measure the image quality of reflective flare removal. Although the bottom line of Fig.\ref{fig8} shows only reflective flare part in the image, both the actual shooting image and the network processing strictly follow the center symmetric effect. Similar to Fig.\ref{fig7}, the image was cropped to better show the details. As the method of obtaining scattering flare dataset is relatively mature, our dataset is designed purely for evaluating the removal of reflective flare. Since the network handles both scattering and reflective flare, it is necessary to remove the scattering flare part to compare the removal ability of reflective flare. As shown in the bottom row of Fig.\ref{fig8}, dataset base and dataset R have almost no effect on reflective flare, all flare still exists. Dataset R, randomly added reflective flare in seems poor and requires reflective flare data guided by centrosymmetry effect to improve the effect. Dataset RP can partly remove reflective flare, but it processes the brightest part poorly, resulting in incorrect black spots. Dataset MR can sense reflective flare, but it processes it poorly overall, causing the image quality to deteriorate. Dataset MRP has the best effect, and can effectively sense and process reflective flare.  Although these data improvements can gradually enhance the network's ability to remove reflective flare, there is currently a lack of reliable datasets with light sources for evaluating this aspect. Since our real shooting datasets can be completely free of any trace of reflective flare, they can serve as reliable and effective test data.

\begin{table*}
  \centering
    \caption{ Quantitative comparison of training on different synthetic data types and testing on different flare removal datasets }
  \begin{tabular}{lccccc}
     \hline
Increase scattering flare   &      &    &    &  \checkmark &\checkmark \\  
Reflective flare in input
                         &       & \checkmark  & \checkmark &\checkmark &\checkmark \\
Scattering flare similarity 
                    &          &      &   &   \checkmark       \\
Reflective flare centrosymmetry &     & & \checkmark  & \checkmark  & \checkmark  \\
 
\hline
Dataset  
  & base &   R &   RP   & MR   & MRP \\
Flare7K unreal (PSNR) 
  & 24.52 (22.6\%)   & 25.46 (10.0\%)   & 24.72 (19.8\%)  & 25.76 (6.29\%)   & \textbf{26.29} (0.00\%)  \\
  
Flare7K unreal (SSIM) 
  &  0.942 (7.41\%)  &   0.944 (3.70\%) &  0.943 (5.56\%) &    \textbf{0.946} (0.00\%)&   0.936 (18.5\%)\\
  
Flare7K real (PSNR)
  & 24.15 (12.7\%) &  24.38 (9.77\%) &  24.43 (9.14\%)  & 24.45 (8.89\%)   & \textbf{25.19} (0.00\%) \\

Flare7K real (SSIM)
  &   0.907  (5.68\%) &   0.902  (11.4\%) &   \textbf{0.912} (0.00\%) &   0.909 (3.41\%) &   \textbf{ 0.912}  (0.00\%)\\
  
Wu's real (PSNR) 
      & 21.07 (36.3\%)   & 21.52 (29.4\%)  & 23.25 (6.05\%)   & 23.37 (4.59\%)   &  \textbf{23.76} (0.00\%) \\

Wu's real (SSIM) 
      &   0.890 (6.80\%) &   0.892 (4.85\%) &   0.896 (0.97\%)&   0.890 (6.80\%) &  \textbf{0.897} (0.00\%)\\
      
Scattering flare real(ours+ Flare7K)(PSNR)  
  & 24.75 (6.66\%)    & 24.82 (5.80\%) & 24.79 (6.17\%)  & 25.09 (2.56\%)  &  \textbf{25.31} (0.00\%)  \\

Scattering flare real(ours+ Flare7K)(SSIM)  
  &   0.916 (6.33\%)  &  0.917 (5.06\%) &   0.919 (2.53\%) &  \textbf{0.921} (0.00\%) &    0.914 (8.86\%) \\
  
Reflective flare real(ours) (PSNR)
  & 33.92 (14.6\%)  & 33.76 (16.7\%)  & 34.16 (11.4\%)   & 27.19 (149\%)  &  \textbf{35.10} (0.00\%) \\ 

Reflective flare real(ours) (SSIM)
  & 0.983 (30.8\%)  &   0.983 (30.8\%) &   0.984 (23.1\%)  &  0.980 (53.9\%) &   \textbf{0.987} (0.00\%)\\ 

Reflective flare real(ours) (LPIPS)
  &  0.1412 (18.6\%) &  0.1384 (16.2\%) & 0.1292  (8.48\%) & 0.1569 (31.7\%) &  \textbf{0.1191} (0.00\%) \\

   \hline
  \end{tabular}

  \label{tab:test_ablation}
\end{table*}
\subsection{Dataset performance evaluation}

In addition to qualitatively comparing the improvements of different data on the network's ability and subjective flare removal results, we also tried to use objective image evaluation indicators to evaluate the effects of different real shooting datasets. The subjective differences between different datasets are shown in Fig.\ref{fig2}. As shown in Table \ref{tab:test_ablation}, the Flare7K unreal dataset has a faster improvement in objective indicators after improvement due to its more consistent construction method with the training dataset, which reflects the effectiveness of our improvement method. However, this effect is not convincing. Unreal is not a real shooting dataset, and the final processing result of the network serves the real world. Therefore, we used the Flare7K real dataset, which has a total of 100 pairs. Compared to the unreal dataset, our data improvement on Flare7K real is more gradual, mainly because there is still a gap between the simulated flare texture and the actual flare. The Flare7K real dataset consists mostly of single light source scenes and a small portion of multiple light source scenes, as shown in Fig.\ref{fig2} (d). This real shooting dataset has scattering flare and very little reflective flare, and the reflective flare is not removed at all.

Based on the shortcomings of existing real-world datasets, we propose our own real-world dataset. Our dataset is divided into two categories: scattering flare and reflective flare. For the scattering flare dataset, we focus more on different color light sources, a larger number of lights, and various light shapes, as shown in Fig.\ref{fig5}. Since the publicly available Flare7K real-world dataset is also valuable, we combine our scattering flare dataset with the Flare7K real-world dataset to form the entire scattering flare real-world dataset. The new scattering flare real-world dataset also reflects the progress of our data improvement methods. However, the progress is less obvious compared to Flare unreal and Flare real datasets. We believe that the light source types in the simulated dataset are still too limited, and it is not adaptable enough to various colors and shapes of light sources, resulting in a smaller improvement relative to the base. The more gradual increase in test result indicators also provides room for the evaluation of future new work, and future improvements can be reflected in our real-world scattered dataset.


For the reflective glare real-world dataset, objective image evaluation only compares the image part of the reflective flare, and the scattering flare part is cropped and removed. Due to the reduced image resolution, the evaluation indicators will be significantly improved, and we only need to focus on the improvement between different data improvement methods. Dataset R, compared to dataset base, randomly adding reflective flare does not improve the ability to remove reflective flare. It is necessary to increase the center symmetry effect, and then dataset RP can improve the ability to remove reflective flare. The subjective image comparison in the bottom row of Fig.\ref{fig8} also shows the same results.  
Dataset MR added multiple scattered light sources and random reflective flare, which can partially remove reflective flare. Our data can show the degradation of reflective flare removal from subjective images and objective indicators with random multi-light sources and non-centrally symmetric reflective flare data. It also shows the advantages of dataset MRP in removing reflective flare.

\section{Conclusion}

Based on the proposed pipeline, most of our flare removal results are satisfactory, and subjectively, some of them are even better than real shooting ground truth. For various real scenes, using our new data generation method can improve subjective and objective indicators. In the manufacture of simulation data, we have proved the similar effect of scattering flare and the central symmetry effect of reflective flare. It is more in line with real scenes than existing datasets. Compared with existing real shooting data, our real shooting dataset generation method has lower cost and more diverse types. We use the characteristics of NeRF to construct a reflection flare dataset, which can accurately process various types of light sources and reflective glare in complex scenes. At the same time, our method can also separately evaluate the network's ability to remove scattering and reflective flare.


However, there are still areas for improvement in single image flare removal. Firstly, when there is strong scattering or reflection in the entire image, the effect is often not good, and texture errors often occur in the central light source area. When there is a lot of flare in the picture or the light source completely covers the background image information, accurate image decomposition methods cannot be perfectly processed due to the lack of essential information. These limitations are mainly caused by the lack of physical information and cannot be solved simply by improving the dataset. Secondly, the current light source scenes are too complex, and we hope to generate more styles of light sources in the simulation dataset rather than simple point source templates.


\nocite{*}

\bibliography{egbib.bib} 

\bibliographystyle{IEEEtran} 

\end{document}